\def\year{2019}\relax
\documentclass[letterpaper]{article} 
\usepackage{aaai19}  
\usepackage{times}  
\usepackage{helvet}  
\usepackage{courier}  
\usepackage{url}  
\usepackage{graphicx}  
\usepackage{color}
\usepackage{mathrsfs}
\usepackage{amsmath}   
\usepackage{subfigure} 
\usepackage{adjustbox}
\begin{document}
%
\title{Hierarchical Photo-Scene Encoder for Album Storytelling} %
\author{Bairui Wang$^1$\thanks{This work was done while Bairui Wang was a Research Intern with Tencent AI Lab.} \qquad Lin Ma$^2$\thanks{Corresponding authors.} \qquad Wei Zhang$^{1\dagger}$ \qquad Wenhao Jiang$^2$ \qquad Feng Zhang$^2$  \\
$^1$School of Control Science and Engineering, Shandong University \qquad $^2$Tencent AI Lab  \\
{\tt\small\{bairuiwong, forest.linma, cswhjiang\}@gmail.com} \\
{\tt\small davidzhang@sdu.edu.cn \qquad jayzhang@tencent.com}
}
\maketitle
\begin{abstract}

In this paper, we propose a novel model with a hierarchical photo-scene encoder and a reconstructor for the task of album storytelling. The photo-scene encoder contains two sub-encoders, namely the photo and scene encoders, which are stacked together and behave hierarchically to fully exploit the structure information of the photos within an album. Specifically, the photo encoder generates semantic representation for each photo while  exploiting temporal relationships among them. The scene encoder, relying on the obtained photo representations, is responsible for detecting the scene changes and generating scene representations. Subsequently, the decoder dynamically and attentively summarizes the encoded photo and scene representations to generate a sequence of album representations, based on which a story consisting of multiple coherent sentences is generated. In order to fully extract the useful semantic information from an album, a reconstructor is employed to reproduce the summarized album representations based on the hidden states of the decoder. The proposed model can be trained in an end-to-end manner, which results in an improved performance over the state-of-the-arts on the public visual storytelling (VIST) dataset. Ablation studies further demonstrate the effectiveness of the proposed hierarchical photo-scene encoder and reconstructor.

\end{abstract}

\section{Introduction}
\begin{figure}
\centering
\includegraphics[width=\hsize]{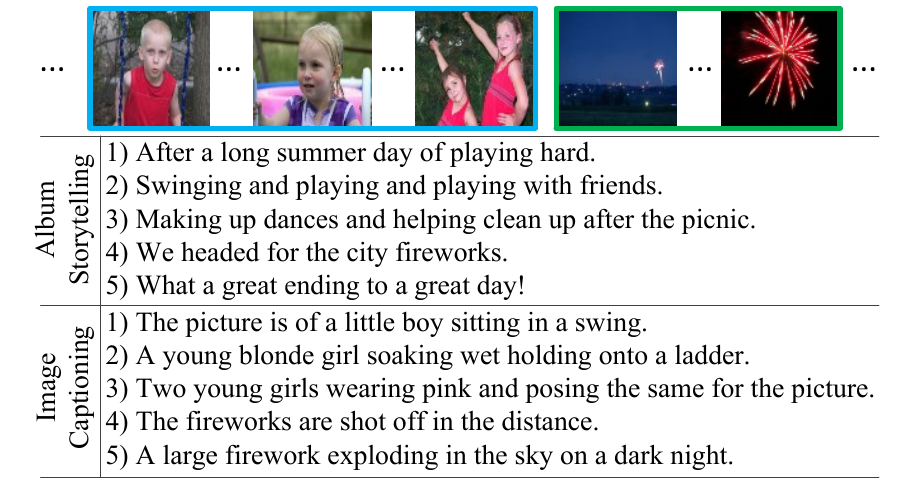}
\caption{Differences between album storytelling and image captioning. Only five representative photos from an album of visual storytelling (VIST)~\cite{huang2016visual} dataset are shown. Sentences in image captioning describe what exactly happens in the current image, while the sentences in album storytelling focus on the sentence coherence and story completeness. Please note that the blue and green boxes represent two different scenes in the album.}
\label{fig:storytell_vs_image_captioning}
\end{figure}

Album storytelling~\cite{yu2017hierarchically,huang2016visual,liu2017video} is a task to  produce a paragraph to describe an ordered photo stream, and has become a hot research topic in the vision and language community. Images in an album are usually redundant and diverse, since people tend to take multiple photos under multiple scenes. To describe an album, the model needs not only to extract the salient contents from the photo stream, but also generate coherent sentences to describe them. Hence, it is totally different from the image captioning task and album storytelling is more challenging.
Human labeled examples of both image captioning and album storytelling are illustrated in Fig.~\ref{fig:storytell_vs_image_captioning}.
In this example, five representative images as well as their labeled captions and story are selected from the album.
It can be observed that the sentences for the image captioning task are independent, only expressing the exact visual content of each image.
On the contrary, the sentences for the album storytelling task take the sentence coherence and story completeness into consideration. Lastly, some sentences in album storytelling might not describe any photos in the stream. The goal of such sentences is to preserve sentence coherence and story completeness. For example, the last sentence in storytelling ``\texttt{what a great ending to a great day!}'' does not describe any images in the album, but it perfectly concludes a story.
For these reasons, we need to consider how to extract related salient information, detect the exiting events or scenes in the album, and finally generate coherent sentences to present the story.

Album storytelling  is usually realized in an encoder-decoder architecture. The encoder relies on the convolutional neural network (CNN) widely used for different works~\cite{zhang2017learning,zhang2018intra,zhang2018long,ma2016learning,qi2018hedging}, to extract the visual feature of each photo and fuses them together to yield the whole album representation.
The decoder usually employs long short-term memory (LSTM) or gated recurrent unit (GRU), to generate the corresponding story.
Liu \textit{et al.}~\cite{liu2017let} bridge the semantically related photos with large visual gap by projecting them into one common semantic space for capturing their visual variance, and construct a coherence matrix to enforce the sentence coherence for storytelling. Yu \textit{et al.}~\cite{yu2017hierarchically} step further by introducing a photo selector between encoder and decoder to automatically choose five photos as the summarization of an album,based on which five sequential sentences are generated as the album story.
For the photos in one album, some of them might reflect events in the same scene, although they may have significant visual variance. For example, in Fig.~\ref{fig:storytell_vs_image_captioning}, the photos highlighted with blue boxes should be in the same scene of ``playing with friends'', while the other two photos highlighted in green boxes should be related to another scene of ``fireworks''. These  scene changes are important for the album storytelling, which are neglected for existing approaches.


To hierarchically exploit the image and scene information, we propose to employ a scene encoder, stacked on the photo encoder,  to detect the scene changes and meanwhile aggregate the scene information. 
Afterwards, the decoder attentively summarizes the photo and  scene representations to form a sequence of album representations and decodes them into multiple coherent sentences. With the scene information taken into consideration, the problem of large visual variances in a photo stream is addressed, which helps improve the sentence coherence in a story.

Additionally, observed the effectiveness of dual learning in machine translation~\cite{tu2017neural} and video captioning~\cite{wang2018reconstruction}, we employ the technique of dual learning to boost the album storytelling performance by
reconstructing the album representations from decoder hidden states.
As such, the hierarchical image and scene information are fully exploited in our model.

The major contributions of this work are summarized as follows:
1) To detect scene changes and aggregate the scene representation, a hierarchical photo-scene encoder for album storytelling is proposed.
2) We propose to reconstruct the attentively aggregated album representations from decoder hidden states, which help exploit the image and scene information. 
3) Extensive results on the video storytelling dataset indicate that the proposed photo-scene encoder and reconstructor can help boost the performance, resulting in a new state-of-the-art on album storytelling.

\section{Related Work}
Album storytelling, a special case of generating natural sentences from visual contents, is related to image captioning~\cite{karpathy2014deep,ma2015multimodal,vinyals2015show,chen2018regularizing,jiang2018learning,jiang2018recurrent} and video captioning~\cite{pan2017video,wang2018reconstruction,wang2018bidirectional,chen2018temporally}, which share some common techniques. In this section, we present a short survey on the related works.


\textbf{Image and Video Captioning.} 
In the early stage, template based methods were proposed to generate captions from images. The sentences are generated by filling a predefined template with contents detected from input image. Later, inspired by the advance in neural machine translation, the encoder-decoder framework~\cite{vinyals2015show} was introduced into image captioning. Nowadays, many variants have been proposed~\cite{xu2015show,he2016dual}.  Recently, reinforcement learning are introduced in this area and achieved remarkable results \cite{rennie2016self,ren2017deep}. Similar to image captioning, encoder-decoder based methods were also proposed for video captioning~\cite{venugopalan2015sequence,pan2016hierarchical}. Different from image captioning, video captioning models need to exploit temporal information in videos, which is the key to boost performance.

\textbf{Album Storytelling.} Different from image and video captioning, the task of album storytelling aims at generating several sentences to describe a set of images which may visual uncorrelated. The first work for this area is \cite{park2015expressing}, in which two datasets named NYC and Disneyland are released. The authors in \cite{park2015expressing} employed a local coherence model \cite{barzilay2008modeling} to parse the patterns of local transitions of sentences in the whole text. After that, Huang \textit{et al.}~\cite{huang2016visual} constructed a dataset named VIST which contains more relevant stories. Liu \textit{et al.} proposed to obtain a semantic space by jointly embedding each image and its corresponding sentence to bridge the images that have similar semantics but large visual variances. Meanwhile, a semantic relation matrix is identified by distance measure in the semantic space, which is used to enforce the sentence coherence \cite{liu2017let}. To automatically summarize the contents of the album for the decoder, Yu \textit{et al.}~\cite{yu2017hierarchically} utilized a learnable selector on the top of visual encoder. Although previous works modeled the relationships between the photos in an album, the effects of scenes are never considered.

\begin{figure*}
\centering
\includegraphics[width=\hsize]{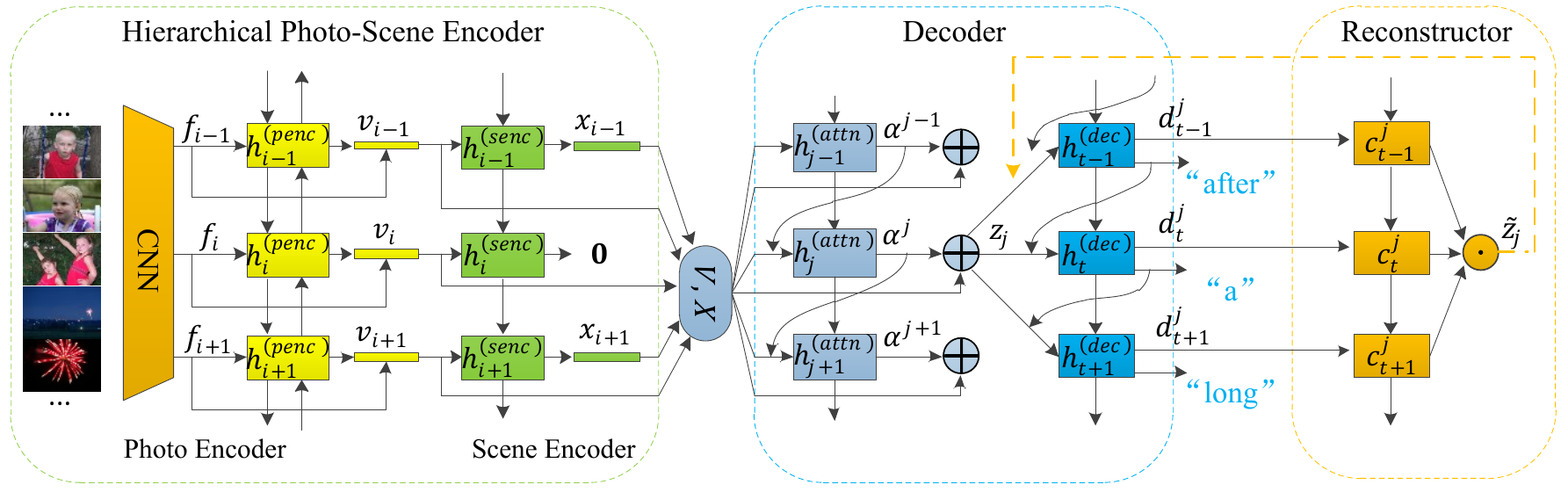}
\caption{The proposed framework for album storytelling. It consists of three components, namely hierarchical photo-scene encoder, decoder and reconstructor. The hierarchical photo-scene encoder is composed of two sub-encoders, namely photo encoder and scene encoder. The  photo encoder extracts the semantic representations of the photos, and the scene encoder explores scene representations.
The decoder attentively summarizes the photo and scene representations and generates multiple coherent sentences as one story for each album. The reconstructor translates story back to the album representations.
Superscripts of hidden states, such as $penc$, $senc$, $attn$, and $dec$, denote photo encoder, scene encoder, attention, and decoder, respectively.
The $\oplus$ and $\odot$ denote weighted sum and average process.}
\label{fig:architecture}
\end{figure*}

\section{Architecture}
For an album with $m$ photos $\mathbf{A}=\left\{\mathbf{a}_1,\mathbf{a}_2, \dots ,\mathbf{a}_m\right\}$, where $\mathbf{a}_i$ denotes the $i$-th photo, the album storytelling aims at generating a story composed of $n$ sentences $\mathbf{S}=\left\{\mathbf{S}_1,\mathbf{S}_2, \dots ,\mathbf{S}_n\right\}$ to describe the album, where $\mathbf{S}_{j} = \left\{s_1^j, s_2^j,\dots, s_{t-1}^j \right\} $ is the $j$-th sentence and $s_{t}^j$ denotes the $i$-th word in sentence $\mathbf{S}_{j}$.
In this paper, we propose an encoder-decoder-reconstructor architecture for the album storytelling, as shown in Fig.~\ref{fig:architecture}.  Specifically, a novel hierarchical photo-scene encoder, containing stacked photo encoder and scene encoder, exploits the hierarchical structure information within the album photos. The decoder dynamically and attentively summarizes the outputs of the photo-scene encoder and decodes several sequential sentences to form a story. A reconstructor that relies on the decoder hidden states is employed to regenerate the summarization by the decoder, which further helps  exploit the information from the album.

\subsection{Hierarchical Photo-Scene Encoder}
The proposed photo-scene encoder contains two sub-encoders, namely photo encoder and scene encoder. The photo encoder models the contents and the temporal information of the album. The scene encoder detects the scene changes. We will present the details in the following subsections.

\subsubsection{Photo Encoder.}

In our model, the image contents are extracted with a CNN, specifically the ResNet~\cite{he2016deep}, and the temporal information in the photo stream is captured with a bidirectional GRU (Bi-GRU). The details of the photo encoder are listed as follows:
\begin{align}
\label{eq:photo_encoder}
\begin{split}
  f_i &= \text{CNN}\left(\mathbf{a}_i\right), \\
  \overrightarrow{h}_i^{(penc)} &= \stackrel{\longrightarrow}{\text{GRU}}\left(f_i, \overrightarrow{h}_{i-1}^{(penc)} \right),  \\
  \overleftarrow{h}_i^{(penc)} &= \stackrel{\longleftarrow}{\text{GRU}}\left(f_i, \overleftarrow{h}_{i-1}^{(penc)} \right),  \\
  v_i &= \text{ReLU}\left(\left[\overrightarrow{h}_i^{(penc)},\overleftarrow{h}_i^{(penc)}\right] + W_f f_i\right),
\end{split}
\end{align}
where $W_f$ is a linear function, $\overrightarrow{h}_i^{(penc)}$ and $\overleftarrow{h}_i^{(penc)}$ are hidden states of Bi-GRU and $v_i$ is the representation for the input photo $\mathbf{a}_i$. Obviously, $v_i$ not only contains the photo content but also captures the context information (other photos) of one album in both forward and backward directions.  In this way, an album $\mathbf{A}$ can be encoded as a sequence of photo representations $\mathbf{V}=\left\{v_1,v_2, \dots ,v_m\right\}$.

\begin{figure}[t]
\centering
\includegraphics[width=\hsize]{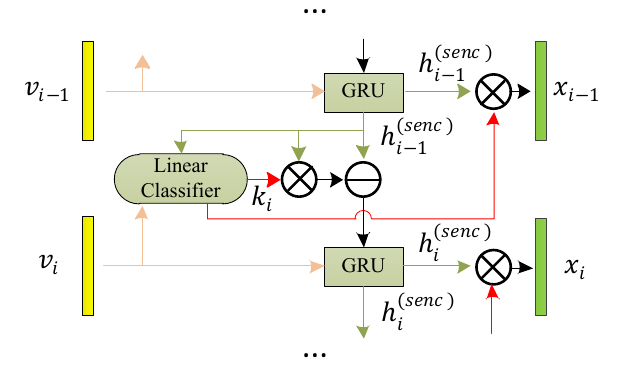}
\caption{The framework of the scene encoder. Taking the phone representations, the scene encoder meanwhile detects the scene changes with a linear classifier and summarizes the scene representations with a GRU when the scene boundaries are detected. The $\otimes$ and $\ominus$ denote multiplication and subtraction process.}
\label{fig:scene_encoder}
\end{figure}


\subsubsection{Scene Encoder.}
\label{sec:scene_encoder}
Different from the videos, in which the visual appearances of adjacent frames are very similar, the photos in an album may be not visually relevant, as illustrated in Fig.~\ref{fig:storytell_vs_image_captioning}. Although these photos are of great differences, they may be taken in the same scene and describe the same activities within an album. In this paper, we identify the semantic discontinuities between photos and thereby detect the scene changes. Meanwhile, each detected scene is further encoded as one scene representation. We adapt a similar boundary detection technique in video~\cite{baraldi2016hierarchical} to detect scene changes in an album based on the obtained photo representations $\mathbf{V}$.
As shown in Fig.~\ref{fig:scene_encoder}, the scene encoder consists of a linear classifier and one GRU to detect scene changes and summarizes the scene information. The two components couple together to generate the final scene representations.  

For a given photo representation sequence, the scene detector acts as a judger to determine whether the current input denotes a start of a new scene, with the consideration of the previous GRU hidden states, which relates to the context scene information. Specifically, the scene detector is realized by a linear classifier: 
\begin{align}
\label{eq:boundary_detector}
\begin{split}
k_i = \left\{
\begin{array}{rcl}
1, & {\textrm{if} \ \sigma\left(w_{sv}^Tv_i + w_{sh}^Th^{(senc)}_{i-1} + b_s\right)>0.5,}\\
0, & {\textrm{otherwise}}
\end{array} \right.
\end{split}
\end{align}
where $k_i$ is the flag indicating whether a new scene is detected, $h^{(senc)}_{i-1}$ denotes the previous hidden state of GRU, $w_{sv}$, $w_{sh}$, and $b_s$ are the learnable parameters and the $\sigma(\cdot)$ denotes a sigmoid function. As the scene detector is a step function, which is a discrete operation, we employ the technique of straight-through estimator~\cite{bengio2013estimating} to back-propagate error signals.


Based on the results of scene detector, GRU updates its previous hidden state $h^{(senc)}_{i-1}$ 
as follow:
\begin{align}
\label{eq:update_state}
  \begin{split}
  	h^{(senc)}_{i-1} &= ( 1 - k_i ) * h^{(senc)}_{i-1}.  
  \end{split}
\end{align}
Therefore, if the scene detector regards the current input $v_i$ as the starting point of a new scene, the flag $k_i$ will be set to 1 and $h^{(senc)}_{i-1}$ will be collected as the final representation of the previous scene. Moreover, as a new scene begins, the hidden state $h^{(senc)}_{i-1}$ 
will be cleared as 0 and the encoding for a new scene begins. If the scene detector does not detect a new scene,  the flag $k_i$ will be 0, and no scene representation needs to be generated.\footnote{Actually, we have to keep the total number of photo and scene representations in the code implementation. So when $k_i=0$, we take $\mathbf{0}$ as a false scene representation and collect it. We also introduced a mask to mark the false and true scene representations.} The hidden state updating rules are the same as in vanilla GRU.


The scene encoder will generate a sequence of scene representations $\mathbf{X}=\left\{x_1,x_2, \dots, x_u\right\}$ for each album, with $x_i$ denoting the hidden state of the GRU when the flag $k_i$ is equal to 1 and $u$ is the number of scenes detected.

\subsection{Decoder}
The obtained  photo and scene representations, \textit{i.e.} $\mathbf{V}$ and $\mathbf{X}$, capture the hierarchical semantic information of the album, which contribute differently to the final story generation.
We combine $\mathbf{V}$ and $\mathbf{X}$ to form a new matrix $\mathbf{R}~=~\left[\mathbf{V}, \mathbf{X}\right]$, and employ attention mechanism to dynamically and attentively summarize the photo and scene representations. We denote the $l$-th column of $\mathbf{R}$ as $r_l$ and denote the sequence of generated summarization as  $\mathbf{Z} = \left\{z_1,z_2,\dots,z_n\right\}$. The procedure of computing  $z_j$ is expressed as:
\begin{align}
	\label{eq:selector_2}
	\begin{split}
		 h_j^{(attn)} &= \text{GRU}\left(\alpha^{j-1}, h_{j-1}^{(attn)}\right), \\
        \tilde{\alpha}^j &= W_\alpha*\tanh\left(W_{\alpha h} h_j^{(attn)} \mathbf{1}^T + W_{\alpha r}\mathbf{R} + b_\alpha\right), \\
        \alpha^j &= \text{softmax} \left( \tilde{\alpha}^j \right), \\
        z_j &= \mathbf{R}\alpha^j,
	\end{split}
\end{align}
where $\mathbf{1}$ is a vector of all ones, and $W_\alpha$, $W_{\alpha h}$, $W_{\alpha r}$ and $b_\alpha$ are learnable parameters.

It can be observed that the attention process on photo and scene representations is relied on GUR. The benefits of such attention strategy lie in two-fold. First, employing attention on both photo and scene representations simultaneously bridges the semantic gaps between each photo and each scene. Second, as the summarizing for current content is affected by the previous attention state, it further enhances the sentence coherence for storytelling.

Based on the generated $n$ album representations $\left\{z_1,z_2,\dots,z_n\right\}$,  $n$ sentences are sequentially generated, composing the final story.
For each album representation $z_j$, we use another GRU to decode its related sentence, which is the same as the decoder in image and video captioning. 
Specifically, GRU takes the album representation $z_j$, the previous word $s_{t-1}^j$, and the hidden state at previous step $h_{t-1}^{(dec)}$ as inputs:
\begin{align}
	\label{eq:decoder_1}
    \begin{split}
        h_t^{(dec)} &= \text{GRU}\left(\left[\text{E}(s_t^j), z_j\right], h_{t-1}^{(dec)}\right), \\
        d_t^j &= \text{MLP}\left(\left[h_t^{(dec)}, z_j\right]\right), \\
   		P\left(s_t^j \mid s_{<t}^j, \mathbf{A}\right) &= \text{softmax}\left(d_t^j\right),
    \end{split}
\end{align}
where $\text{E}(\cdot)$ is a word embedding function that turns a word into a learnable vector.
The output state $h_t^{(dec)}$ is concatenated with the album representation $z_j$, which generates the word distribution with another $\text{MLP}$ and $\text{softmax}$ function. $P$ denotes the word probability for word $s_t^j$ of $j$-th sentence at time step $t$ when the generated partial caption $s_{<t}^j$ (\textit{i.e.} $\left\{s_1^j, s_2^j,\dots, s_{t-1}^j \right\}$) is known.


\subsection{Reconstructor}
On top of the decoder, we build a GRU-based reconstructor to reconstruct the generated album representations $\mathbf{Z}$ based on the decoder hidden states.
As such, the information from the sentences to the album can be further exploited, which is believed to benefit the album storytelling.

As shown in Fig.~\ref{fig:architecture},
the logits $\mathbf{D}_j=\left\{d_1^j, d_2^j, \dots, d_n^j \right\}$ for the $j$-th sentence contains the sentence semantic information. The reconstructor first performs the mean pooling on $\mathbf{D}_j$ to obtain the global sentence information $\bar{d}^j=\frac{1}{n}\sum_{i=1}^{n}d_i^j$. Then at each time step, GRU is used to reconstruct the corresponding album representation:
\begin{align}
\label{eq:recons_2}
 c_t^j = \text{GRU}\left(\left[d_t^j, \bar{d}^j\right], c_{t-1}^j\right),
\end{align}
where $c_t^j$ is reconstructor hidden state of the $j$-th sentence.
Here we use $\mathbf{C}_j = \left\{c_1^j, c_2^j, \dots, c_n^j\right\}$ to represent the hidden states of the reconstructor. Finally, we obtain the reconstructed album representation by averaging $\mathbf{C}_j$ to obtain $\tilde{z}_j$, which will be used to
compare with ${z}_j$ to obtain reconstruction loss.

The reconstructor we design in this paper is different from \cite{wang2018reconstruction} on twofold. First, we reproduce one album representation with all hidden states from the decoder after a sentence is generated, while the model in \cite{wang2018reconstruction} needs to reconstruct the feature of each frame. This is mainly attributed to the differences between storytelling and captioning tasks. Second, what we reconstruct is the attentively summarized album representations, which contains the photo and scene information as well as their temporal relationships. In contrast, only video frame features are considered to be reconstructed in~\cite{wang2018reconstruction}.

\subsection{Loss Function and Training Strategy}
In this subsection, we present the loss function used at each step and introduce the training strategy.


Given the albums, we aim at minimizing the negative log probability of the story sentences in the decoder step:
\begin{align}
	\label{eq:optim}
    \begin{split}
        \mathcal{L}_{dec}(\theta) = \sum_{y=1}^{N}\sum_{j=1}^{n} -\log P\left(\mathbf{S}^y_j \mid \mathbf{A}^y\right),
    \end{split}
\end{align}
where $N$ denotes the total number of albums and the number $n$ is the number of sentences in the story for an album. The sentences $\mathbf{S}_j$ are generated word by word, with the probability defined  as:
\begin{align}
	\label{eq:probability}
    P\left(\mathbf{S}_j \mid \mathbf{A} \right) = \prod_{t=1}^{T} P\left(s_t^j \mid s_{<t}^j, \mathbf{A}\right),
\end{align}
where $T$ denotes the length of sentence.

In order to capture the temporal coherence in an album,
 we follow Yu \textit{et al.}~\cite{yu2017hierarchically} to employ the order-preserving constraint, and the loss function is expressed as:
\begin{align}
	\label{eq:rank_loss}
    \begin{split}
    &\mathcal{L}_{rank}(\theta) =\\
    & \sum_{y=1}^{N} \sum_{j=1}^n \max\left(0, 1-\log P\left({{\mathbf{S}}'}_j^y \mid \mathbf{A}^y\right) +\log P\left(\mathbf{S}_j^y \mid \mathbf{A}^y\right) \right).
    \end{split}
\end{align}
Sentences from a story are shuffled to obtain negative instances ${\mathbf{S}}'$.



For the reconstructor, the Euclidean distance between reconstructed and original album representations are regarded as the reconstruction loss:
\begin{align}
	\label{eq:recons_loss}
    \mathcal{L}_{rec}(\theta_{rec}) = \sum_{y=1}^{N}\sum_{j=1}^{n} \left\| \tilde{z}_j^y - z_j^y \right\|^{2}
\end{align}
Considering the encoder, decoder, and reconstructor together, the complete loss for training our model is defined as:
\begin{align}
\label{eq:final_loss_2}
    \mathcal{L}(\theta, \theta_{rec}) = \mathcal{L}_{dec}(\theta) + \lambda \mathcal{L}_{rank}(\theta) + \mu \mathcal{L}_{rec}(\theta_{rec}),
\end{align}
where $\lambda$ and $\mu$ are the trade-off parameters. To train the model, we train the encoder and decoder first. Then the parameters of encoder and the album summarization part of the decoder are fixed, the reconstructor is trained.


\section{Experiments}

In this section, we evaluate the effectiveness of our proposed model on album storytelling. We first describe the datasets used for evaluation, followed by a brief description of competitor models. Afterward, the experimental results on album storytelling are illustrated and discussed.

\begin{table*}
\begin{center}
\caption{ Performance comparisons with different competitor models on the testing set of the VIST dataset in terms of BLEU~\cite{papineni2002bleu}, CIDEr~\cite{vedantam2015cider}, METEOR~\cite{banerjee2005meteor}, and ROUGE-L~\cite{lin2004rouge} scores (\%).
The scores of the competitor baselines, namely enc-dec, enc-attn-dec, h-attn, and h-attn-rank  are directly copied  from~\cite{yu2017hierarchically} for fair comparisons. '-' indicates the unreported score.}
\label{table:results1}
\begin{adjustbox}{width=0.7\textwidth}
\begin{tabular}{c|c|c|c|c|c|c|c}
\hline
models &BLEU-1  &BLEU-2  & BLEU-3  &BLEU-4 & CIDEr & METEOR  & ROUGE-L \\ \hline \hline
enc-dec & - & - & 19.58 & - & 4.65  & 33.02  &29.23  \\
enc-attn-dec  & - & - & 19.73 & - & 4.96&  32.98  & 28.94  \\
h-attn  & - & - & 20.53  & - & 6.84  & 33.81  & 29.82  \\
h-attn-rank  & - & - & 20.78  & - & 7.38 & 33.94  & 29.82  \\ \hline
HP  & 61.22 & 37.58  & 21.31 & 12.08  & 7.44 & 34.16 & 29.73 \\
HPS   & 61.79 & 37.61 & 21.39  & 12.10 & 7.75 & 34.23  & 29.91 \\
HPR   & 61.83 & 37.72 & 21.39  & 12.09 & 7.61 & 34.35  & 29.79  \\
HPSR  & \textbf{61.94} & \textbf{37.82} & \textbf{21.51}  & \textbf{12.21} & \textbf{8.03}  & \textbf{34.43}  & \textbf{31.17}  \\ \hline
\end{tabular}
\end{adjustbox}
\end{center}
\end{table*}

\subsection{Datasets}
To compare with existing methods, we evaluate the proposed album storytelling model on the visual storytelling dataset (VIST)~\cite{huang2016visual}, which is particularly created for the task of album storytelling.
Specifically, VIST consists of about 10K albums with about 200K unique photos. Each album is described with 5 stories, with each story containing 5 sequential and coherent sentences. Moreover, 5 photos are selected with order from each album as its corresponding summary.  The VIST dataset is split into three parts, \textit{i.e.}, $8,031$ albums for training, $998$ for validation, and $1,011$ for testing.




\subsection{Implementation Details}
In this section, we describe the detailed configurations and implementation details of our proposed whole network, including the hierarchical photo-scene encoder, decoder, and reconstructor.

For the sentences, the word that occurs less than 5 times are eliminated.
And each sentence within each story is truncated to 25 words, with each word is embedded as a 512-dimensional vector.


For album, same as~\cite{yu2017hierarchically}, we also truncate the photo stream, which contains only 40 photos, instead of using only 5 labeled photos for each album. For each photo, we use the ResNet101 pre-trained on the ILSVRC-2012-CLS dataset~\cite{russakovsky2015imagenet} as the feature extractor to generate 2048-dimensional feature.
The sizes of all GRUs in the hierarchical model and linear function in both photo and scene encoders are set as 512.
For decoder, since the number of scene representations are dynamic, the dimension of weight vector is decided by the total number of photo and scene features. The hidden states of GRUs are initialized to zero, except that the attention GRU is initialized by the final state of photo encoder.

We use the Adam~\cite{kingma2014adam} as the optimizer, with the initial learning rate being set as 0.0004 while other parameters using the recommended parameters. The training process terminates when the value of CIDEr metric on validation stops growing in 30 validations.
Training the whole network performs in two stages. First, the encode-decoder is trained until convergence. Afterwards, the reconstructor is stacked to perform a joint training with the loss function defined in Eq.~\eqref{eq:final_loss_2}.

\subsection{Competitor Models}


In this subsection, we mainly compare our method with the competitor models in~\cite{yu2017hierarchically}, as we aim to generate stories on the whole album, instead of several manually selected photos as in~\cite{wang2018no,liu2017let}.
\begin{itemize}
\item enc-dec: a seq2seq model with the encoder and decoder realized in RNN. The encoder encodes all photos sequentially and the decoder decodes the last hidden state of the encoder into one story.
\item enc-att-dec: an attention model sharing the similar encoder-decoder architecture with enc-dec. An attention mechanism performs on all hidden states of the encoder to dynamically summarize a representation for the story decoding.
\item h-attn: a model with hierarchical encoder-selector-decoder framework, where the selector chooses 5 photos to summarize the content of album, based on which decoder generates 5 sequential sentences and aggregates them as one complete story.
\item h-att-rank: this model has the identical hierarchical architecture with the one in h-attn but considers the ranking loss defined in Eq.~\eqref{eq:rank_loss}.
\end{itemize}

To reveal the impact of each component, we also provide the results of our models by removing certain components. The variants of our method are listed as follows.
\begin{itemize}
\item HP: the model only contains the photo encoder and decoder, which acts as the base architecture of our proposed method.
\item HPS: based on HP, the model incorporates the scene encoder to build the hierarchical photo-scene encoder.
\item HPR: based on HP, the reconstructor is stacked on the decoder.
\item HPSR: the proposed complete model includes the hierarchical photo-scene encoder, decoder, and reconstructor.
\end{itemize}

\subsection{Experimental Results and Analysis}
In this subsection, we first examine the contributions of the proposed scene encoder and reconstructor, and then demonstrate the effectiveness of the whole model. It should be noted that the weight $\lambda$ for ranking loss in Eq.~\eqref{eq:final_loss_2} is set as 0.2, same as that in h-attn-rank, which ensures fair comparisons. The weight $\mu$ for reconstructor is set as 0.8. Experiments about different trade-off parameters will be discussed in the following.
Compared to the best baseline model h-attn-rank, HP gets better results on CIDEr, BLEU-3, and METEOR, which demonstrate that the attention mechanism in HP indeed summarizes the proper album representations for decoder to generate sentences. By incorporating the photo-scene encoder and reconstructor, the performances can be consistently improved, yielding  superior performances of HPS and HPR over HP.
The improvement can be attributed to the following two reasons. On one hand, the proposed hierarchical photo-scene encoder in HPS effectively captures the temporal information in photo stream, and  exploits the scene semantic information from albums. Such hierarchical exploited information can help to well characterize the album representations and thereby  benefit the story generation. On the other hand, the reconstructor in HPR exploits dual information by reconstructing album representations, and thereby further enhances the performance of storytelling model.


By considering the the hierarchical photo-scene encoder and the reconstructor together, HPSR sets a new state-of-the-art performance, demonstrating the best performance in terms of all metrics. Therefore, the hierarchical photo-scene encoder and reconstructor can not only improve the performance individually but also cooperate together to further improve our proposed storytelling model.

\begin{figure*}[!ht]
\centering
\includegraphics[width=\hsize]{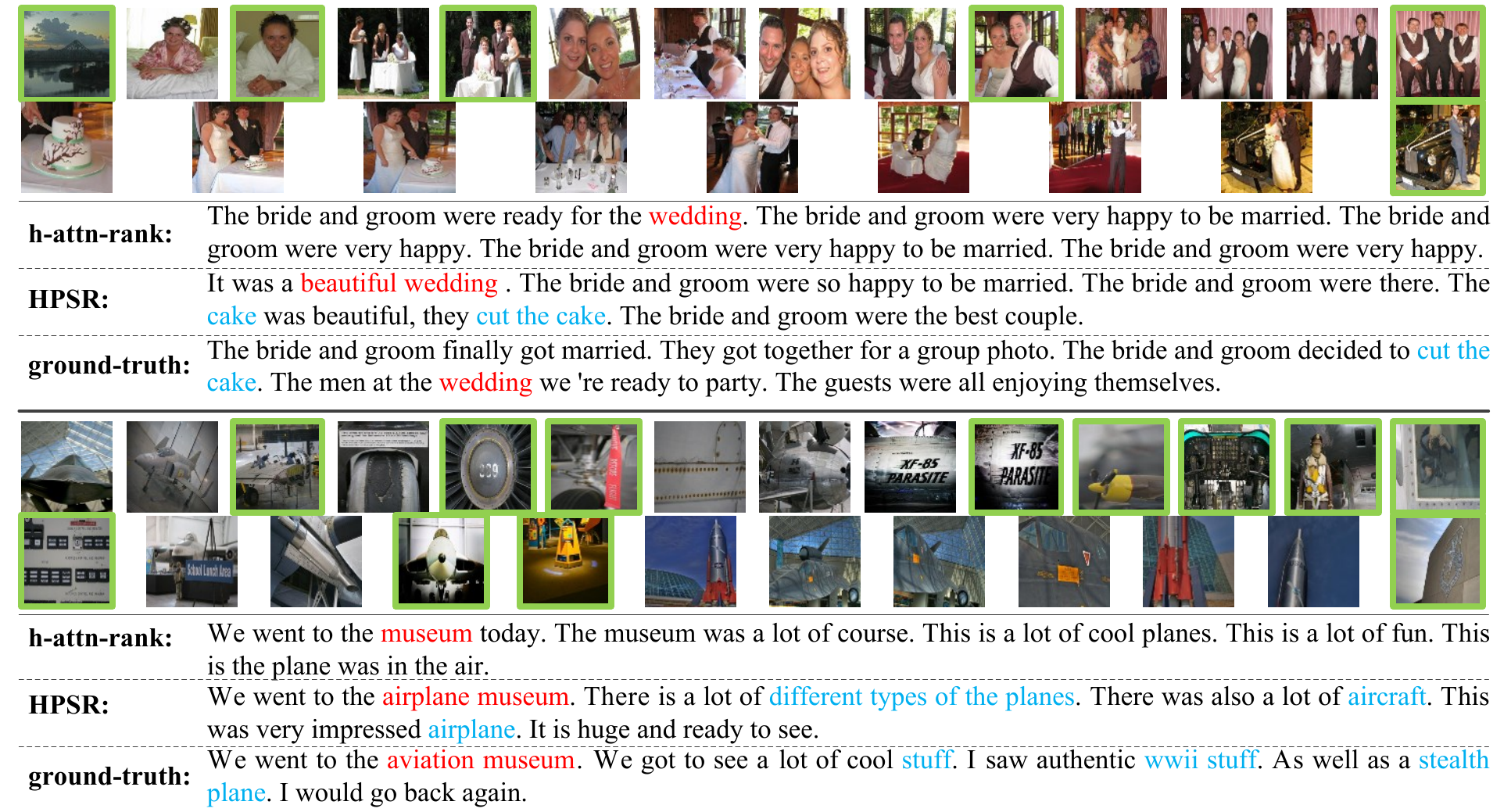}
\caption{Some story examples on the VIST dataset generated by h-attn-rank and HPSR. Due to the page limit, one of the five ground-truth stories is shown. Words related with story themes are in red and blue. And scene boundaries detected are shown with green border.}
\label{fig:results}
\end{figure*}

\subsubsection{Qualitative Analysis.}
Some qualitative examples are illustrated in Fig.~\ref{fig:results}.
In the first sample, h-attn-rank expresses the same meaning within the generated sentences, which is ``the bride and groom are happy''. For our HPSR model,  the sentence indicates that it is a \textit{'wedding'} (in red) in the beginning, and then mentions the cake-cutting (in blue).
In the end, the story expresses praise to the bride and groom, which is more similar to a story by human.
Compared with the story generated by h-attn-rank, HPSR generates more coherent sentences and yields a complete story.
In the second sample, the story generated by HPSR tells us the property of museum \textit{'airplane museum'}(in red). And HPSR generates  different words for describing the aircrafts, such as \textit{'plane'}, \textit{'aircraft'} and \textit{'airplane'} (in blue), although it fails to understand the relationship of them. In contrast, h-attn-rank only tell us just \textit{'planes'} in \textit{'museum'}.

Moreover, we use green box to highlight the detected scene boundaries in Fig.~\ref{fig:results}, We can accurately detect the scene changes from outdoor to indoor and from single to crowd in the first example, as well as the scene changes from aircrafts to buildings in the second example. It clearly demonstrates that the scene encoder can identify the scene and then utilize and aggregate the scene information for further boosting the performances.
Actually, detecting scene changes is affected by huge visual variances between photos. For example, scene boundaries in the first sample are clearer than those in the second one, as photos in the first album changes more smoothly than the second album. In future, we will focus on how to exploit the scene information from a deeper semantic level.

\subsubsection{Effects of Trade-off Parameters.}
The hyper-parameters $\lambda$ and $\mu$ in Eq.~\eqref{eq:final_loss_2} balance the contributions of the  ranking loss and reconstruction loss. In this subsection, we study the effects of these two parameters. Experimental results illustrate that our model is robust to the trade-off parameters.

First,  $\lambda$ in HPS varies from 0.0 to 0.5 with step of 0.1 to increase the role of ranking loss with $\mu$ being set as 0.0 to exclude the influence of reconstruction loss.
Results are shown in Table~\ref{table:results_lambda}. Note that the results with $\lambda=0.2$  correspond to those of HPS in Table~\ref{table:results1}. It can be observed that scores obtained with ranking loss are better than those of model without considering ranking loss ($\lambda=0$ ). It demonstrates that ranking loss is beneficial to generate plausible stories. However, paying more attention on ranking loss may not help. When $\lambda$ is larger than 0.2, scores on all metrics will decrease.

\begin{table}[h]
\begin{center}
\caption{Performance of HPS when $\mu=0$ with different values of $\lambda$ on the testing set of VIST (\%). 
}
\begin{adjustbox}{width=0.345\textwidth} %
\label{table:results_lambda}
\begin{tabular}{c|c|c|c|c}
\hline
$\lambda$    & BLEU-3 & CIDEr & METEOR  & ROUGE-L \\ \hline
0.0   &20.27 &6.99  & 33.55 & 29.59 \\
0.1  &20.90 &7.33 &33.78 &29.71 \\
0.2   &\textbf{21.39} &\textbf{7.75}  &\textbf{34.23}  &\textbf{29.91}  \\
0.3  &21.31 &7.71 &34.12 &29.90 \\
0.4   &21.31 &7.64 &34.06  &29.88  \\
0.5  &21.24 &7.60 &34.02 &29.84 \\
\hline
\end{tabular}
\end{adjustbox}
\end{center}
\end{table}

Second, in HPSR, we fix $\lambda$ as 0.2 and increase the trade-off parameters $\mu$ from 0.0 to 1.0 with step of 0.2 to examine the contributions of reconstruction loss.
As shown in Table~\ref{table:results_mu}, performances can be always improved by introducing the reconstruction loss.
The dual information can be more comprehensively exploited by reconstructing attentively summarized album representations from the decoder hidden states. Thus the visual storytelling performance can be improved.
In this paper, $\mu$ is set as 0.8 according to the experimental results in Table~\ref{table:results_mu}.

\begin{table}[h]
\begin{center}
\caption{Performance of HPSR when $\lambda=0.2$ with different values of $\mu$ on the testing set of VIST (\%). 
}
\label{table:results_mu}
\begin{adjustbox}{width=0.345\textwidth}
\begin{tabular}{c|c|c|c|c}
\hline
$\mu$    & BLEU-3 & CIDEr & METEOR  & ROUGE-L \\ \hline
0.0   &21.39 &7.75 &34.23 &29.91 \\
0.2   &21.40 &7.92 &34.35 &29.72 \\
0.4   &21.41 &7.84 &34.41 &30.75 \\
0.6  &21.46 &7.87 &34.36 &30.78 \\
0.8  &\textbf{21.52} &\textbf{8.03} &\textbf{34.43} &\textbf{31.17} \\
1.0  &21.49 &7.91 &34.31 &30.02 \\
\hline
\end{tabular}
\end{adjustbox}
\end{center}
\end{table}

\section{Conclusions}
In this work, we proposed a novel network with a hierarchical photo-scene encoder and a reconstructor for the task of album storytelling, which exploits the hierarchical visual and scene semantic information within an album and the dual information between the album and story, respectively.
Jointly trained by minimizing the negative log probabilities of the generated words and maximizing the similarity of the original and reconstructed album representations, the proposed model archives the state-of-the-art performance on the VIST dataset, which indicates the superiority of our proposed hierarchical photo-scene encoder and reconstructor on generating coherent sentence for describing the album.

\section*{Acknowledgments}
This work was supported by the National Key Research and Development Plan of China under Grant 2017YFB1300205, NSFC Grant no. 61573222, Major Research Program of Shandong Province 2018CXGC1503, and Fundamental Research Funds of Shandong University 2016JC014.

\bibliographystyle{aaai}
\small
\bibliography{aaai}
\end{document}